\documentclass[runningheads]{llncs}
\usepackage{graphicx}

\usepackage{tikz}
\usepackage{comment}
\usepackage{amsmath,amssymb} 
\usepackage{color}
\usepackage{subcaption}
\usepackage{sidecap}
\usepackage{caption}
\usepackage{bbm} 
\usepackage{gensymb}
\usepackage{multirow}
\usepackage{url}            
\newcommand{\etal}{\textit{et al}.}
\newcommand{\ie}{\textit{i}.\textit{e}.}
\newcommand{\eg}{\textit{e}.\textit{g}.}
\newcommand*{\affmark}[1][*]{\textsuperscript{#1}}

\usepackage[accsupp]{axessibility}  

\begin{document}
\pagestyle{headings}
\mainmatter
\def\ECCVSubNumber{2861}  

\title{A Unified Framework for Domain Adaptive Pose Estimation} 

%
\author{Donghyun Kim\inst{2}\thanks{Equal Contribution.} \and Kaihong Wang\inst{1}$^\star$ \and \\ Kate Saenko\inst{1,2} \and Margrit Betke\inst{1} \and Stan Sclaroff\inst{1} }
\institute{\affmark[1]Image and Video Computing, Department of Computer Science, Boston University, \affmark[2]MIT-IBM Watson AI Lab\\
	\email{\{{donhk, kaiwkh, saenko, betke, sclaroff\}@bu.edu}}}

\authorrunning{D. Kim and K. Wang \etal}
%
\maketitle

\begin{abstract}
While pose estimation is an important computer vision task, it requires expensive annotation and suffers from domain shift. In this paper, we investigate the problem of domain adaptive 2D pose estimation that transfers knowledge learned on a synthetic source domain to a target domain without supervision. While several domain adaptive pose estimation models have been proposed recently, they are not generic but only focus on either human pose or animal pose estimation, and thus their effectiveness is 
somewhat 
limited to specific scenarios. 
In this work, we propose a unified framework that generalizes well on various domain adaptive pose estimation problems. We propose to align representations using both input-level and output-level cues (pixels and pose labels, respectively), which facilitates the knowledge transfer from the source domain to the unlabeled target domain. Our experiments show that our method achieves state-of-the-art performance under various domain shifts. Our method outperforms existing baselines on human pose estimation by up to 4.5 percent points (pp), hand pose estimation by up to 7.4 pp, and animal pose estimation by up to 4.8 pp for dogs and 3.3 pp for sheep. These results suggest that our method is able to mitigate domain shift on diverse tasks and even unseen domains and objects (\eg, trained on horse and tested on dog). Our code will be publicly available at: \url{https://github.com/VisionLearningGroup/UDA_PoseEstimation}.

\end{abstract}

\section{Introduction}

Recent developments in dense prediction tasks, \eg, semantic segmentation~\cite{BadrinarayananKeCi17,ChenPaKoMuYu18,LongSeDa15,RonnebergerFiBr15} or pose estimation~\cite{NewellYaDe16,SunXiLiWa19,XiaoWuWe18}, are limited by the difficulty in the acquisition of massive datasets~\cite{CordtsOmRaRenEeBrFrRoSc16,DengDoSoLiLi009,GeigerLeStUr13,IonescuPaOlSm14} due to the expensiveness as well as the unreliability that originates from the annotation phase. 
In addition, these models often perform poorly under domain shift. In this work, we address the problem of 2D pose estimation in the unsupervised domain adaptation (UDA) setting. 
The UDA setting allows us to train a pose estimation model with supervision from synthetic (source) domains, where data and accurate annotations are much cheaper to acquire, and optimize the model's performance on an unlabeled real (target) domain. Nevertheless, the domain gap between  
source and 
target domains due to 
distributional shift greatly undermines the ability of the model to transfer learned knowledge across different domains.  This is a challenge that has been addressed previously for UDA for classificational tasks   
~\cite{HoffmanWaYuDa16,LiYuVa19,LongCaWaJo15,SaitoWaUsHa18}.

\begin{figure}[t]
     \centering
     \begin{subfigure}[b]{0.28\textwidth}
         \centering
        \begin{subfigure}[t]{0.47\textwidth}
            \includegraphics[width=0.99\linewidth]{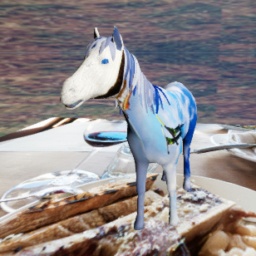}  
        \end{subfigure}
        \begin{subfigure}[t]{0.47\textwidth}
            \includegraphics[width=0.99\linewidth]{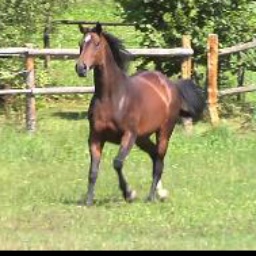}   
        \end{subfigure}
        \par\medskip
        \begin{subfigure}[t]{0.47\textwidth}
            \includegraphics[width=0.99\linewidth]{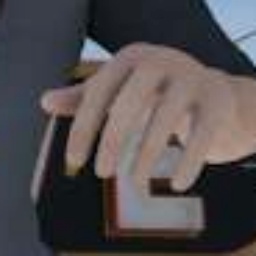}  
        \end{subfigure}
        \begin{subfigure}[t]{0.47\textwidth}
            \includegraphics[width=0.99\linewidth]{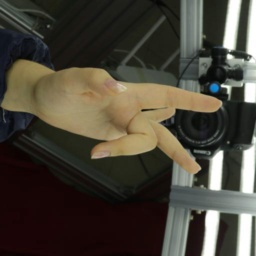}   
        \end{subfigure}
        \par\medskip
        \begin{subfigure}[t]{0.47\textwidth}
            \includegraphics[width=0.99\linewidth]{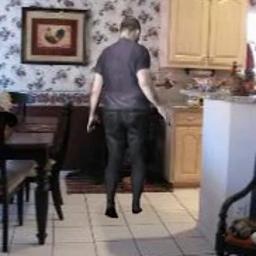}  
            \captionsetup{labelformat=empty}
            \caption{Synthetic}
        \end{subfigure}
        \begin{subfigure}[t]{0.47\textwidth}
            \includegraphics[width=0.99\linewidth]{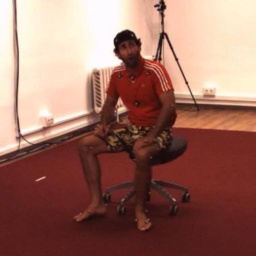}   
            \captionsetup{labelformat=empty}
            \caption{Real}
        \end{subfigure}
     \end{subfigure}
     \hfill
     \begin{subfigure}[b]{0.69\textwidth}
         \centering
        \begin{subfigure}[t]{0.19\textwidth}
            \includegraphics[width=0.99\linewidth]{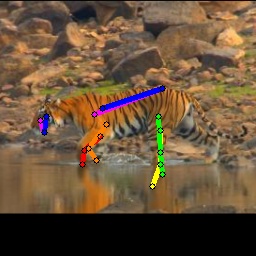}   
        \end{subfigure}
        \begin{subfigure}[t]{0.19\textwidth}
            \includegraphics[width=0.99\linewidth]{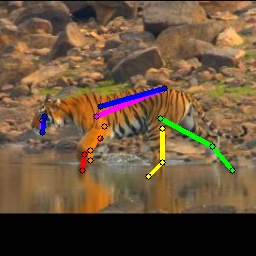}  
        \end{subfigure}
        \begin{subfigure}[t]{0.19\textwidth}
            \includegraphics[width=0.99\linewidth]{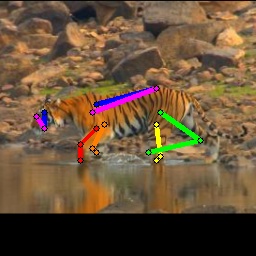}  
        \end{subfigure}
        \begin{subfigure}[t]{0.19\textwidth}
            \includegraphics[width=0.99\linewidth]{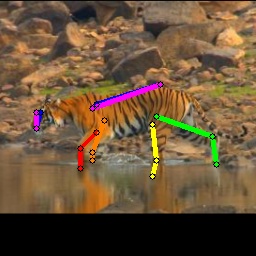}  
        \end{subfigure}
        \begin{subfigure}[t]{0.19\textwidth}
            \includegraphics[width=0.99\linewidth]{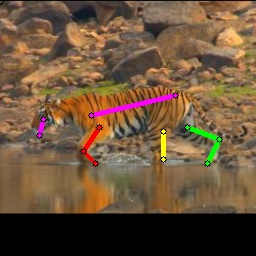}  
        \end{subfigure}
        \par\medskip
        \begin{subfigure}[t]{0.19\textwidth}
            \includegraphics[width=0.99\linewidth]{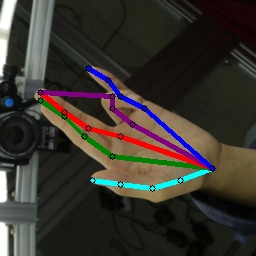}   
        \end{subfigure}
        \begin{subfigure}[t]{0.19\textwidth}
            \includegraphics[width=0.99\linewidth]{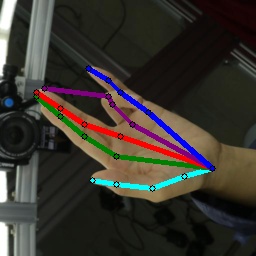}  
        \end{subfigure}
        \begin{subfigure}[t]{0.19\textwidth}
            \includegraphics[width=0.99\linewidth]{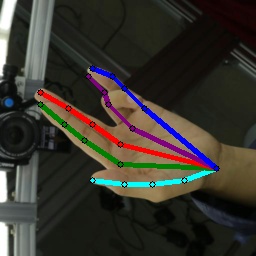}  
        \end{subfigure}
        \begin{subfigure}[t]{0.19\textwidth}
            \includegraphics[width=0.99\linewidth]{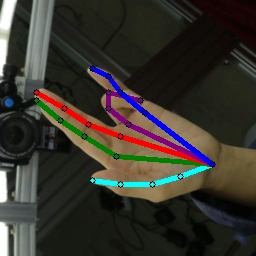}  
        \end{subfigure}
        \begin{subfigure}[t]{0.19\textwidth}
            \includegraphics[width=0.99\linewidth]{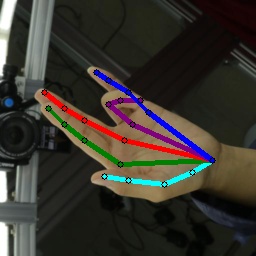}  
        \end{subfigure}
        \par\medskip
        \begin{subfigure}[t]{0.19\textwidth}
            \includegraphics[width=0.99\linewidth]{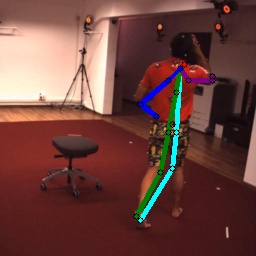}   
            \captionsetup{labelformat=empty}
            \caption{CCSSL}
        \end{subfigure}
        \begin{subfigure}[t]{0.19\textwidth}
            \includegraphics[width=0.99\linewidth]{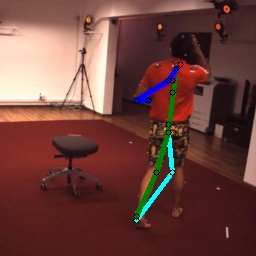}  
            \captionsetup{labelformat=empty}
            \caption{UDA-A}
        \end{subfigure}
        \begin{subfigure}[t]{0.19\textwidth}
            \includegraphics[width=0.99\linewidth]{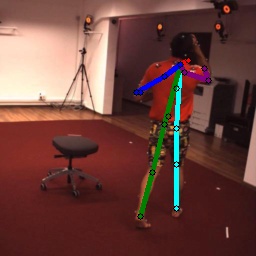}  
            \captionsetup{labelformat=empty}
            \caption{RegDA}
        \end{subfigure}
        \begin{subfigure}[t]{0.19\textwidth}
            \includegraphics[width=0.99\linewidth]{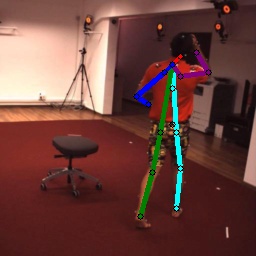}  
            \captionsetup{labelformat=empty}
            \caption{Ours}
        \end{subfigure}
        \begin{subfigure}[t]{0.19\textwidth}
            \includegraphics[width=0.99\linewidth]{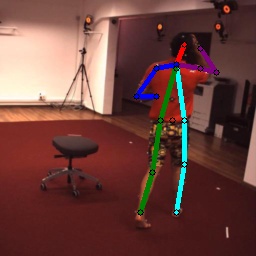}  
            \captionsetup{labelformat=empty}
            \caption{GT}
        \end{subfigure}
     \end{subfigure}
     (a) \hspace*{5.5cm} (b) \hspace*{2.5cm}
    \caption{(a) Top row: An example of high input-level variance 
    in animal pose estimation benchmarks (large color and textual differences). 
    Middle and bottom row: An example of high output-level variance in human and hand pose estimation benchmarks (large pose differences). (b) Visualization of pose estimation results from baselines, our method and ground-truth (GT). Note that both CCSSL and UDA-Animal(UDA-A) are proposed for animal pose estimation, while RegDA is only validated on hand and human pose estimation tasks. Most baseline methods suffer from performance degradation when applied to the other task. In comparison, our unified framework can more accurately estimate poses of hand, human and animal under various scenarios}
    \label{fig:fig-1}
\end{figure}

Less attention has been paid to using UDA for regression tasks such as 2D pose estimation. Existing works are not generic but specifically target human pose estimation (RegDA~\cite{JiangJiWaLi0aLo21}) or animal pose estimation 
(CCSSL~\cite{MuQiHaYu20}, UDA-Animal~\cite{LiLe21}). 
A reason for this specialization may be the nature of the particular datasets used in those benchmarks. Animal datasets typically show large input-level variance (Fig.~\ref{fig:fig-1}-(a)top) while human and hand datasets show large output-level variance (Fig.~\ref{fig:fig-1}-(a)middle and bottom). 
Therefore, existing UDA methods do not generalize well 
to different objects of interest, for example, training and testing a human pose estimation model on an animal species or vice versa.

To address the aforementioned problems and keep the framework model-agnostic, we propose to bridge the domain gap via both input-level and output-level adaptations, \ie, alignments across domains in both the input and the output space of a pose estimation model.
In input-level adaptation, we first translate images through a pre-trained style transfer model~\cite{HuangBe17} that can extract similar visual features and bridge the gap between domains. 
In output-level adaptation, we borrow the architecture of Mean Teacher~\cite{DFrenchMaFi18,TarvainenVa17} that enforces consistency in the output space of a student and a teacher model to generate reliable pseudo labels and learn from the unlabeled target domain. 

As a typical approach for pose estimation, heatmap regression~\cite{tompson2014joint} predicts probabilities of the presence of keypoints in 2D space. However, unlike the output probabilities from other classification models that represent relative significance in the output space and sum to 1, the output heatmaps from a pose estimation model, which learns the task as predicting absolute value, are not normalized. 
The learning objectives of the student model, guided by the non-normalized output from the teacher model, will then be diverted from learning relative significance in the heatmap to learning absolute values, which is a more challenging task as the output space is no longer constrained. 
Therefore, the stability of the consistency learning is greatly undermined, and the lack of constraints leads to a problem we identify as a drift effect.
Meanwhile, the drifted output heatmap also poses challenges while selecting confident guidance from the teacher model via the confidence thresholding method in Mean Teacher, as it potentially brings in noise that further deteriorates unsupervised learning. Therefore, we propose to normalize the output of the teacher model to make the guidance more stable. 
Our empirical results demonstrate the importance of this simple yet crucial step to deploy the Mean Teacher model for regression tasks.

In addition to revising consistency learning for the regression task, we design differing self-guiding strategies for student and teacher,
developed especially for domain adaptive pose estimation. With style transfer, we generate target-like images from the source images and train a model to minimize the supervised loss with source labels. For the target domain, we generate source-like images from the target images to generate high-quality pseudo-labels from the teacher and give better guidance to the student model. In addition, in the student branch, we adaptively apply an occlusion mechanism, which has shown promising effectiveness especially in pose estimation tasks~\cite{DevriesTa17,KeChQiLy18,XieWaZeWa21}, based on the feedback of the teacher model. This strengthens the robustness of the pose estimation model.

In experiments we validate the effectiveness and generalization ability of our method under various scenarios including hand and human pose estimation as well as animal pose estimation. Our results show significant improvements over the existing domain adaptive pose estimation baselines by up to 4.5 percent point (pp) on hand pose,  7.4 pp on human pose estimation, and 4.8 pp for dog as well as 3.3 pp for sheep on animal pose estimation. Additionally, we present generalization experiments where we test models on unseen datasets or categories (\ie, different animals), and verify the 
generalization capability. Further sensitivity analysis and ablation studies reveal the relation and interaction between modules and explain the effectiveness of each component of our unified framework.
To summarize, our contributions in this work include:
\begin{itemize}
  \item Unlike prior works, we propose a unified framework 
  for general pose estimation that generalizes well on diverse objects in the pose estimation task.
  \item We propose a multi-level (\ie, input-level and output-level) alignment method for domain adaptive pose estimation that can effectively address domain gap problems in different levels under different scenarios (\eg, Fig.~\ref{fig:fig-1}-(a)). 
  \item We address the drifting problem in the Mean Teacher paradigm and facilitate its learning from unlabeled data especially for pose estimation tasks.
  \item We unified benchmarks from human pose estimation and animal pose estimation in this work and present state-of-the-art performance in general pose estimation, providing a stronger baseline in this line of research.
\end{itemize}

\section{Related Works}

\subsection{Pose Estimation}

Pose estimation has become an active research topic for years. In this paper, we focus on 2D pose estimation. Hourglass~\cite{NewellYaDe16} is one of the dominant approaches for human pose estimation which applies an encoder-decoder style network with residual modules and finally generate heatmaps. 
A mean-squared error loss is applied between the predicted heatmap and ground-truth heatmap consisting of a 2D Gaussian
centered on the annotated joint location~\cite{tompson2014joint}.  Xiao \etal~\cite{XiaoWuWe18} propose a simple baseline model that combines upsampling and deconvolutional layers without using residual modules. HRNet~\cite{SunXiLiWa19} is proposed to maintain high-resolution in the model and achieves promising results.
In this paper, we adopt the architecture of the Simple baseline model~\cite{XiaoWuWe18} following~\cite{JiangJiWaLi0aLo21} to fairly compare our method with prior domain adaptation algorithms.

\subsection{Unsupervised Domain Adaptation}

Unsupervised Domain Adaptation (UDA) aims to bridge the domain gap between a labeled source domain and unlabeled target domain. Existing domain adaptation methods utilize adversarial learning~\cite{ganin2017domain,long2018conditional}, minimize feature distances using MMD~\cite{gretton2012optimal}, optimal transport~\cite{bhushan2018deepjdot}, pixel-level adaptation~\cite{HoffmanTzPaZhIsSaEfDa18}, or maximum classifier discrepancy~\cite{SaitoWaUsHa18} for classification. In addition several other UDA methods have been proposed for dense prediction tasks including semantic segmentation~\cite{HoffmanWaYuDa16,LiYuVa19,TsaiHuScSoYaCh18,YangSo20} and depth estimation~\cite{KunduLaRa19,KunduUpPaBa18,RodriguezMi20}. Compared to other visual tasks, domain adaptation for regression tasks are still not well explored. 

\subsection{Domain Adaptive Pose Estimation}

There are two categories in domain adaptation pose estimation: (1) For human pose estimation, RegDA~\cite{JiangJiWaLi0aLo21} made changes in MDD~\cite{zhang2019bridging} for human and hand pose estimation tasks, which measures discrepancy by 
estimating false predictions on the target domain. (2) For animal pose estimation, pseudo-labeling based approaches have been proposed in~\cite{LiLe21,MuQiHaYu20}. Mu~\etal~\cite{MuQiHaYu20} proposed invariance and equivariance consistency learning with respect to transformations as well as temporal consistency learning with a video. Li~\etal~\cite{LiLe21} proposed a refinement module and a self-feedback loop to obtain reliable pseudo labels.
Besides, WS-CDA~\cite{CaoTaFaShTaLu19} leverages human pose data and a partially annotated animal pose dataset to perform semi-supervised domain adaptation.
In our experiments, we observed that (1) and (2) do not work well on the other tasks. A likely cause could be that each estimation task has different types of domain shifts, as shown in Fig~\ref{fig:fig-1}(a). To address this, we propose a unified framework that generalizes well on diverse tasks by utilizing both input-level and out-level cues.

\section{Method}
\subsection{Preliminaries}

\begin{figure}[t]
\centering
    \includegraphics[width=0.95\linewidth]{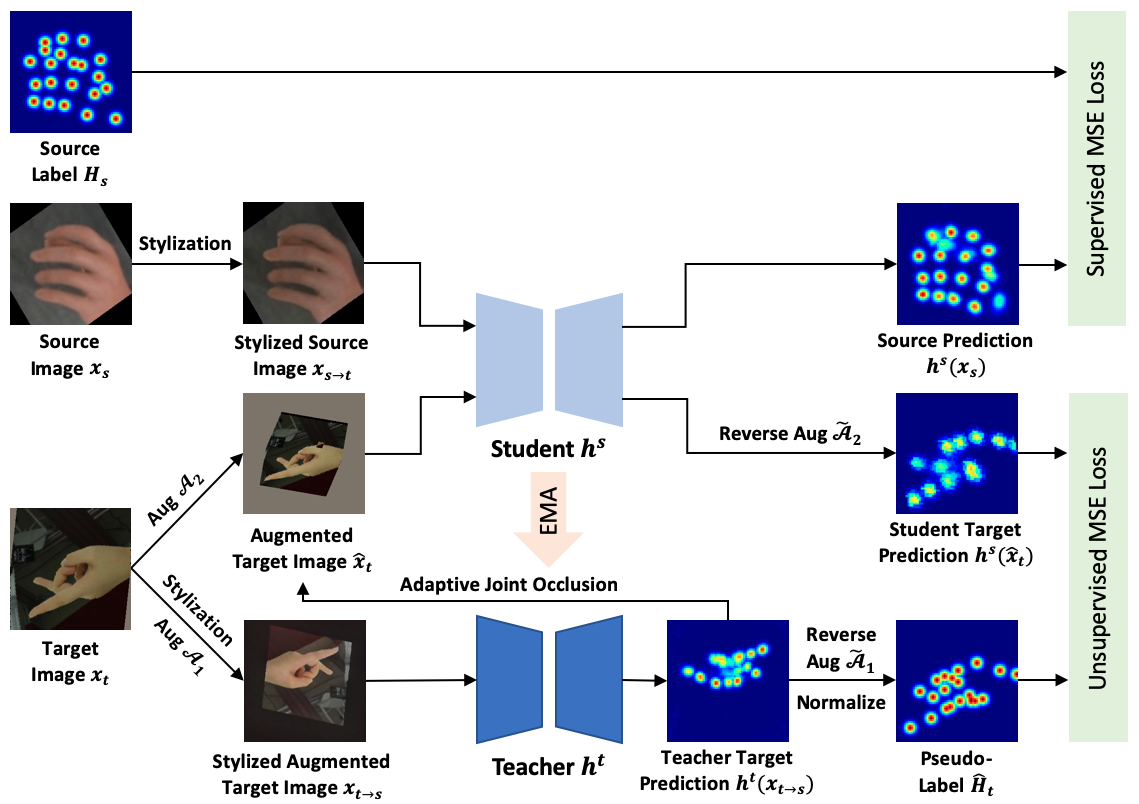}   
    \caption{An overview of our unified framework comprising a supervised branch that learns from source domain data with corresponding annotation, as well as an unsupervised branch that learns from unlabeled target domain data. We perform domain alignment both in the input-level via style-transfer with style references from the opposite domain, and the output-level of the model that guides the training on the target domain with more reliable pseudo-labels. The student model is trained by the combination of two losses, while the teacher model is updated with the exponential moving average weights of the student}
    \label{fig:pipeline}
\end{figure}

Given a labeled pose dataset $\mathcal{S}= \{(x_s^i,y_s^i)\}_{i=1}^N$ in source domain consisting of $N$ pairs of images $x_s \in \mathbb{R}^{H \times W \times 3} $ and corresponding annotation heatmap $y_s \in \mathbb{R}^{K \times 2}$ representing the coordinates of $K$ keypoints, as well as an unlabeled pose dataset $\mathcal{T} =\{x_t^i\}_{i=1}^M$ in target domain consisting of $M$ images $x_t \in \mathbb{R}^{H \times W \times 3}$, we aim to learn a 2D pose estimation model $h$ and optimize the performance on the target domain. 
Typically, the pose estimation model $h$ is pre-trained on the source domain dataset in a supervised manner to learn pose estimation from heatmaps $H_s=L(y_s)$, where $H \in \mathbb{R}^{K\times H'\times W'}$ with the output heatmap size $H'$ and $W'$, generated through the heatmap generating function $L:\mathbb{R}^{K\times2}\rightarrow\mathbb{R}^{K\times H'\times W'}$, with classic MSE loss: $L_{sup}=\frac{1}{N} \sum_{x_s\in \mathcal{S}} ||h(x_s)-H_s||_2$.


\subsection{Input-level Alignment via Style Transfer}

Different from prior works~\cite{HoffmanTzPaZhIsSaEfDa18,HoffmanWaYuDa16,TzengHoSaDa17} that adopt adversarial learning, we propose to perform input-level alignments via style transfer for the sake of efficiency and simplicity. We borrow notations from AdaIN~\cite{HuangBe17} and follow its settings and training procedure to extract content features from a content image $c$ and style feature from a style image $s$ through a pre-trained VGG~\cite{SimonyanZi14a} model $f$. Formally, style transfer is performed with a generator $g$ pre-trained as in AdaIN:

\begin{equation} \label{eq:style-transfer}
T(c,s,\alpha)=g(\alpha t+(1-\alpha)f(c))
\end{equation}
where $t=\text{AdaIN}(f(c),f(s))$ is the combination of content and style feature through adaptive instance normalization and $\alpha$ is the content-style trade-off parameter. Exemplar results are illustrated in the appendix. With a fixed AdaIN model, we transform source domain images with styles from target domain $x_{s \rightarrow t} = T(x_s,x_t,\alpha)$ and revise the supervised loss above:

\begin{equation} \label{eq:supervised_revised}
L_{sup}=\frac{1}{N} \sum_{x_s\in \mathcal{S}} ||h(x_{s \rightarrow t})-H_s||_2
\end{equation}

\subsection{Output-level Alignment via Mean Teacher}

To better exploit information from the unlabeled target domain, we adopt the paradigm of Mean Teacher that trains a student pose estimation model $h^s$ by the guidance produced by its self-ensemble, i.e., the teacher pose estimation model $h^t$ in an unsupervised learning branch. The input image for each model is augmented by $\mathcal{A}_1$ and $\mathcal{A}_2$ stochastically sampled from data augmentation $\mathcal{A}$. While the student $h^s$ is updated according to the  supervised loss in Eq.~\ref{eq:supervised_revised} and self-guidance from the teacher $h^t$, the weights of the latter are updated as the estimated moving average of the former. 

On the opposite direction to the supervised learning branch that transforms the source image to the target domain, we also propose to transform the target domain image back to the direction of the source domain where supervised learning happens and bridge the domain gap when generating guidance from the teacher model. Formally, we take a source domain image as the style reference and generate $x_{t\rightarrow s}=T(\mathcal{A}_1(x_t), x_s, \alpha)$. After that, we pass the transformed image through the teacher model and get corresponding heatmap $H_t=h^t(x_{t\rightarrow s})$.

\begin{figure}[t]
\centering
    \begin{subfigure}[t]{0.47\textwidth}
        \includegraphics[width=1.0\linewidth]{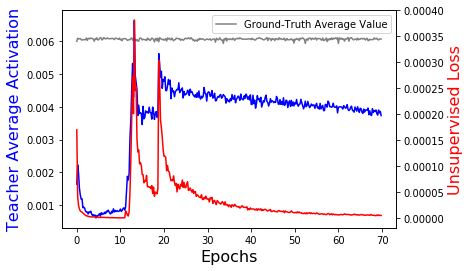}   
    \end{subfigure}
    \begin{subfigure}[t]{0.4\textwidth}
        \includegraphics[width=1.0\linewidth]{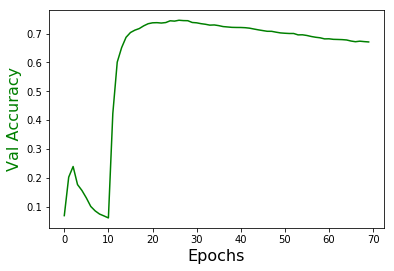}  
    \end{subfigure}
    \caption{Drift effect and its influence to the consistency learning. In the left plot, the gray curve represents the averaged value of the ground-truth heatmap. We observe that the averaged activation of teacher's output (blue curve) gradually decreases and drift away from the gray curve while minimizing the unsupervised loss (red curve). This leads to a degradation in accuracy as shown in the right plot}
    
    \label{fig:fig-drift}
\end{figure}

With the generated guidance heatmap from the teacher model, we still need to address the drifting effect that brings in instability in the unsupervised learning, as illustrated in Fig.~\ref{fig:fig-drift}. 
Technically, we generate pseudo-labels $\hat{H_t}=L(\hat{y_t})$ with the positions that produce maximum activation $\hat{y_t}=\arg\max_pH_{t}^{:,p}$ from each keypoints of the guidance heatmap to normalize the heatmap. We also revise the typical thresholding mechanism using a fixed value in Mean Teacher and determine the confidence threshold $\tau_{conf}$ with the top $p\%$-th values among maximum activation from each keypoint to exclude noises and further improve the quality of the self-guidance.

In addition to improving the quality of the teacher's prediction, we also seek to challenge the student model by adaptively occluding the input to the student model according to feedback from the teacher. To be more specific, we mask the regions where the teacher model makes confident prediction of a keypoint with activation greater than $\tau_{occ}$ via an occlusion operation: $\hat{x}_t=O(\mathcal{A}_2(x_t),\tau_{occ})$, and let the student to learn robust prediction based on its contextual correlation with other keypoints from teacher's pseudo-label after reversing augmentations $\Tilde{\mathcal{A}}_1$ and $\Tilde{\mathcal{A}}_2$.
Overall, the student model $h^s$ will be guided by the normalized heatmap $\hat{H_t}$ via an unsupervised learning loss on keypoints $k$ producing maximum activation $H_t^{k,\hat{y}_t}$ greater than or euqal to threshold $\tau_{conf}$:

\begin{equation} \label{eq:unsupervised}
L_{unsup}=\frac{1}{M}  \sum_{x_t\in \mathcal{T}} \sum_{k=0}^{K} \mathbbm{1}(H_t^{k,\hat{y}_t}\ge \tau_{conf})  ||\Tilde{\mathcal{A}}_1(\hat{H_t}^{k})-\Tilde{\mathcal{A}}_2(h^s(\hat{x}_t)^{k})||_2
\end{equation}

Combining our supervised learning loss from Eq.~\ref{eq:supervised_revised} and unsupervised learning loss from Eq.~\ref{eq:unsupervised}, we present the illustration for the overall pipeline in Fig.~\ref{fig:pipeline} and the final learning objectives:

\begin{equation} \label{eq:overall}
L=L_{sup} + \lambda L_{unsup}
\end{equation}

\section{Experiments}

To verify the effectiveness and reliability of our method under various pose estimation scenarios (hand, human body, animals), we conducted experiments on benchmark datasets in those domains (Sec.~\ref{exp:dataset}) and 
compared our methods with SOTA baselines (Sec.~\ref{exp:comparison}).
We also evaluated our method on domain generalization tasks where we tested our models on unseen domains (\ie, different datasets) and objects (\ie, different animals) (Sec.~\ref{exp:generalization}).
Finally, we present a sensitivity analysis on hyper-parameters and ablation studies to analyze the contribution and interaction between each component in our paradigm (Secs.~\ref{exp:sensitivity_analysis} and~\ref{exp:ablation}). 

\subsection{Experiment Protocols}
\label{exp:protocol}
We adopted the architecture of Simple Baseline~\cite{XiaoWuWe18} as our pose estimation model for both $h^s$ and $h^t$, with backbone of pre-trained ResNet101~\cite{HeZhReSu16}. Following Simple Baseline and RegDA, we adopted Adam~\cite{KingmaBa14} as the optimizer and set the base learning rate as 1e-4. It decreased to 1e-5 at 45 epochs and 1e-6 at 60 epochs, while the whole training procedure consisted of 70 epochs. The batch size was set to 32 and there are in total 500 iterations for each epoch. 
The confidence thresholding ratio $p$ is 0.5, while the occlusion thresholding value $\tau_{occ}$ is 0.9. The momentum $\eta$ for the update of the teacher model is 0.999 and the unsupervised learning weight was set to 1 to balance the supervised and unsupervised loss to a similar level.
Also, the model was only trained by the supervised loss on the source domain for the first 40 epochs. 
On the basis of augmentation in RegDA, we added 
rotation (-30\degree, 30\degree) and random 2D translation (-5\%, 5\%) for the input source and target domain images. Finally, it should be noted that we used the same hyper-parameters for all experiments, did not tune the number of training epochs on test sets, and always report the accuracy of models from the last epoch. 
As for the architecture and optimization procedure of the style transfer model, we follow settings in AdaIN, except that we pre-train the model bidirectionally, i.e., both source and target domain image can be a content or a style image. Additional details can be found in the appendix.

\subsection{Dataset}
\label{exp:dataset}
\textbf{Rendered Hand Pose Dataset}~\cite{ZimmermannBr17} (RHD) provides $44k$ synthetic hand images including $41.2k$ training images and $2.7k$ test images along with corresponding 21 hand keypoints annotations. \textbf{Hand-3D-Studio}~\cite{ZhaoWaXiWa20} (H3D) is a real-world multi-view indoor hand pose images dataset with $22k$ frames. We follow RegDA's policy to split $3.2k$ frames as the test set. \textbf{FreiHAND}~\cite{ZimmermannCeYaRuArBr19} includes $44k$ frames of real-world multi-vew hand pose images with more varied pose and view points. It contains $130k$ training image, and we still follow settings in RegDA to select $32k$ test images. \textbf{SURREAL}~\cite{VarolRoMaMaBlLaSc17} provides more than 6 million synthetic human body pose images with annotations. \noindent\textbf{Human3.6M}~\cite{IonescuPaOlSm14} contains 3.6 million frames of real-world indoor human body pose images captured from videos. We follow protocols in~\cite{LiCh14} and split 5 subjects (S1, S5, S6, S7, S8) as the training set and 2 subjects (S9, S11) as test set. \textbf{Leeds Sports Pose}~\cite{JohnsonEv10} (LSP) is a real-world outdoor human body pose dataset containing $2k$ images. \textbf{Synthetic Animal Dataset}~\cite{MuQiHaYu20} is a synthetic animal pose dataset rendered from CAD models. The dataset contains 5 animal classes, horse, tiger, sheep, hound, and elephant, each with $10k$ images. \textbf{TigDog Dataset}~\cite{PeroRiSuFe15} includes $30k$ frames from real-world videos of horses and tigers. \textbf{Animal-Pose Dataset}~\cite{CaoTaFaShTaLu19} provides $6.1k$ real-world images from 5 animals including dog, cat, cow, sheep, and horse.

\subsection{Experimental Results}
\label{exp:comparison}

\noindent\textbf{Baselines.} We consider the following SOTA baselines: semi-supervised learning based CCSSL~\cite{MuQiHaYu20}, UDA-Animal~\cite{LiLe21}, and RegDA~\cite{JiangJiWaLi0aLo21} under various adaptation tasks. 
For the sake of fair comparison, we re-train CCSSL and UDA-Animal with the backbone of ResNet-101 as ours, and train CCSSL jointly among all categories in animal pose estimation tasks. Oracle is the performance of a model trained jointly with target 2D annotations following previous works  

\noindent\textbf{Metrics.} We adopt the evaluation metric of Percentage of Correct Keypoint (PCK) for all experiments and report PCK@0.05 that measures the ratio of correct prediction within the range of 5\% with respect to the image size.

\begin{SCtable}[][t]

\centering
\setlength{\tabcolsep}{4pt}
\resizebox{0.5\textwidth}{!}{\caption{\label{tab:r2h2} Prediction accuracy PCK@0.05 on \textsl{RHD$\rightarrow$H3D}, i.e., source dataset is RHD, target dataset H3D, for four hand parts and the full hand. Higher values are better
}
\begin{tabular}{c | c c c c c} 
 \hline \hline
 Method & MCP & PIP & DIP & Fin & All \\ 
 \hline
 Source only & 67.4 & 64.2 & 63.3 & 54.8  & 61.8\\ 

 Oracle  & 97.7 & 97.2 & 95.7 & 92.5 & 95.8\\
 \hline
 CCSSL~\cite{MuQiHaYu20} & 81.5 & 79.9 & 74.4 & 64.0 & 75.1 \\

 UDA-Animal~\cite{LiLe21}  & 82.3 & 79.6 & 72.3 & 61.5 & 74.1\\

 RegDA~\cite{JiangJiWaLi0aLo21} & 79.6 & 74.4 & 71.2 & 62.9 & 72.5 \\ 
 \hline
 Ours  & \textbf{86.7} & \textbf{84.6} & \textbf{78.9} & \textbf{68.1} & \textbf{79.6} \\ 
 \hline
\end{tabular} }

\end{SCtable}

\noindent\textbf{Results on Hand Pose Estimation.} First, we present the adaption results on the hand pose estimation task \textsl{RHD$\rightarrow$H3D} on 21 keypoints. We report different anatomical parts of a hand including metacarpophalangeal (MCP), proximal interphalangeal (PIP), distal interphalangeal (DIP), and fingertip (Fin). 
Our baselines can greatly improve the performance of their pose estimation model
on the target domain (Table~\ref{tab:r2h2}), while UDA-Animal, which is originally proposed for animal pose estimation tasks, achieves a performance of 75.1\%. In comparison, our method outperforms all the baseline methods by a noticeable margin of 4.5\% and reaches 79.6\%.

\noindent\textbf{Results on Human Body Pose Estimation.} As for the adaptation in human body pose estimation, we measure the performance of all baselines and ours in the task of \textsl{SURREAL$\rightarrow$Human3.6M} and \textsl{SURREAL$\rightarrow$LSP} on 16 keypoints on the human body grouped with different parts, \ie, shoulders, elbow, wrist, hip, knee, and ankle. RegDA can successfully adapt its model closer to the target domain, while CCSSL and UDA-Animal, designed for animal pose estimation, fail to adapt under such scenarios (Table~\ref{tab:s2h&l}). This could probably be because their self-guidance paradigm is more hyper-parameter sensitive and cannot guarantee to generalize to other scenarios, including the high out-level variance (\ie, high pose variance) in human pose estimation. Our method, in contrast, enables effective and robust unsupervised learning via the heatmap normalization which addresses the drift effect and therefore ensures the high quality of the self-guidance.

\begin{table}[t!]
\centering
\caption{ PCK@0.05 on \textsl{SURREAL$\rightarrow$Human3.6M} and \textsl{SURREAL$\rightarrow$LSP}. Sld: Shoulder, Elb: Elbow}
\setlength{\tabcolsep}{0.8pt}
\fontsize{7.5}{10} \selectfont
\begin{tabular}{c | c c c c c c c | c c c c c c c} 
 \hline \hline
 \multirow{2}{*}{Method} & \multicolumn{7}{c|}{\textsl{SURREAL$\rightarrow$Human3.6M}} & \multicolumn{7}{c}{\textsl{SURREAL$\rightarrow$LSP}} \\
  & Sld & Elb & Wrist & Hip & Knee & Ankle & All & Sld & Elb & Wrist & Hip & Knee & Ankle & All \\ 
 \hline
 Source only & 69.4 & 75.4 & 66.4 & 37.9 & 77.3 & 77.7 & 67.3 & 51.5 & 65.0 & 62.9 & 68.0 & 68.7 & 67.4 & 63.9\\ 

 Oracle & 95.3 & 91.8 & 86.9 & 95.6 & 94.1 & 93.6 & 92.9 & - & - & - & - & - & - & - \\
 \hline
 CCSSL~\cite{MuQiHaYu20}  &  44.3 & 68.5 & 55.2 & 22.2 & 62.3 & 57.8 & 51.7  & 36.8 & 66.3 & 63.9 & 59.6 & 67.3 & 70.4 & 60.7\\

 UDA-Animal~\cite{LiLe21} & 51.7 & 83.1 & 68.9 & 17.7 & 79.4 & 76.6 & 62.9 & 61.4 & 77.7 & 75.5 & 65.8 & 76.7 & 78.3 & 69.2\\

 RegDA~\cite{JiangJiWaLi0aLo21} & 73.3 & 86.4 & 72.8 & \textbf{54.8} & 82.0 & 84.4 & 75.6 & 62.7 & 76.7 & 71.1 & 81.0 & 80.3 & 75.3 & 74.6\\ 
 \hline
 Ours & \textbf{78.1} & \textbf{89.6} & \textbf{81.1} & 52.6 & \textbf{85.3} & \textbf{87.1} & \textbf{79.0} & \textbf{69.2} & \textbf{84.9} & \textbf{83.3} & \textbf{85.5} & \textbf{84.7} & \textbf{84.3} & \textbf{82.0}\\ 
 \hline
\end{tabular} 
\label{tab:s2h&l}
\end{table}

\noindent\textbf{Results on Animal Pose Estimation.} We finally compare our method with the baselines in domain adaptive animal pose estimation under \textsl{SynAnimal}$\rightarrow$\textsl{Tig\-Dog} and \textsl{SynAnimal$\rightarrow$AnimalPose} as shown in Tables~\ref{tab:animal} and~\ref{tab:animal_other}. In \textsl{SynAnimal}$\rightarrow$\textsl{Tig\-Dog}, we follow settings in UDA-Animal and estimate 18 keypoints from different parts including eye, chin, shoulder, hip, elbow, knee, and hoof of horse and tiger shared in the Synthetic Animal and the TigDog datasets. In \textsl{SynAnimal$\rightarrow$AnimalPose}, we also perform adaptation on the hound and sheep categories for 14 keypoint estimation of eye, hoof, knee, and elbow. For a fair comparison, we run all experiments with the same data augmentation as in CCSSL and UDA-Animal for all tasks, as these augmentations provide crucial improvement (see first and second rows in  Table~\ref{tab:animal}). The first row in Table~\ref{tab:animal} represents the reported~\cite{LiLe21}  
source-only performance without augmentations; the second row 
with augmentation, which, e.g., increases the performance from 32.8\% to 71.4\% in the horse keypoint estimation (column All).

Among the baseline methods, UDA-Animal achieves the best performance in estimating a horse's pose and approaches the oracle performance from a model trained jointly by the annotated source and target domains. Our method achieves slightly lower performance in the horse set that is close to the oracle level but slightly outperforms UDA-Animal in the tiger set. 

In despite of the promising results in \textsl{SynAnimal$\rightarrow$TigDog}, we observe that UDA-Animal significantly underperforms than RegDA and ours in the AnimalPose dataset from Table~\ref{tab:animal_other}. This is because  \textsl{SynAnimal}$\rightarrow${\em Ani\-mal\-Pose} is more challenging than \textsl{SynAnimal$\rightarrow$TigDog} by comparing the accuracy of source only models (32.2\% vs. 71.4\%).  Even though we can still see improvements from the source only with augmentations, CCSSL and UDA-Animal face more noisy pseudo-labels during self-training possibly due to their hyper-parameter sensitivity, so that improvements are marginal. On the contrary, RegDA shows noticeable improvement compared to source only. Our method can handle these challenging settings via heatmap normalization in pseudo-labeling and obtain the best performance in these experiments in both categories.

\begin{table}[t!]
\centering
\setlength{\tabcolsep}{0.8pt}
\fontsize{7}{10} \selectfont
\caption{ PCK@0.05 on \textsl{SynAnimal$\rightarrow$TigDog}. Sld: shoulder, Elb: Elbow. Source only$^*$ indicates training on only source domain data with strong augmentation}
\resizebox{\textwidth}{!}{\begin{tabular}{c | c c c c c c c c | c c c c c c c c} 
 \hline \hline
 \multirow{2}{*}{Method} & \multicolumn{8}{c|}{Horse} & \multicolumn{8}{c}{Tiger} \\
 \cline{2-17}
  & Eye & Chin & Sld & Hip & Elb & Knee & Hoof & All & Eye & Chin & Sld & Hip & Elb & Knee & Hoof & All \\ 
 \hline
 Source only & 49.3 & 53.5 & 31.3 & 53.5 & 38.7 & 28.7 & 18.3 & 32.8 & 42.8 & 32.1 & 24.2 & 51.1 & 32.6 & 28.1 & 32.7 & 33.2\\ 
 Source only$^*$ & 87.1 & 91.4 & 69.4 & 76.3 & 70.1 & 71.3 & 61.9 & 71.4 & 91.1 & 86.5 & 46.5 & 67.9 & 44.3 & 53.1 & 63.2 & 60.7\\ 
 Oracle & 92.0 & 95.8 & 73.6 & 90.9 & 84.4 & 84.2 & 79.1 & 84.1 & 98.5 & 97.4 & 75.1 & 94.7 & 74.1 & 76.0 & 81.6 & 82.1\\ 
 \hline
 CCSSL~\cite{MuQiHaYu20}  & 89.3 & \textbf{92.6} & 69.5 & 78.1 & 70.0 & 73.1 & 65.0 & 73.1 & 94.3 & 91.3 & 49.5 & 70.2 & 53.9 & 59.1 & 70.2 & 66.7\\

 UDA-Animal~\cite{LiLe21} & 86.9 & 93.7 & \textbf{76.4} & \textbf{81.9} & 70.6 & \textbf{79.1} & \textbf{72.6} & \textbf{77.5} & 98.4 & 87.2 & 49.4 & \textbf{74.9} & 49.8 & 62.0 & 73.4 & 67.7\\

 RegDA~\cite{JiangJiWaLi0aLo21} & 89.2 & 92.3 & 70.5 & 77.5 & 71.5 & 72.7 & 63.2 & 73.2 & 93.3 & 92.8 & 50.3 & 67.8 & 50.2 & 55.4 & 60.7 & 61.8\\ 
 \hline
 Ours & \textbf{91.3} & 92.5 & 74.0 & 74.2 & \textbf{75.8} & 77.0 & 66.6 & 76.4 & \textbf{98.5} & \textbf{96.9} & \textbf{56.2} & 63.7 & \textbf{52.3} & \textbf{62.8} & \textbf{72.8} & \textbf{67.9}\\ 
 \hline
\end{tabular} }
\label{tab:animal}
\end{table}

\begin{table}[t!]
\centering
\caption{  PCK@0.05 on \textsl{SynAnimal$\rightarrow$AnimalPose}. Source only$^*$ indicates training on only source domain data with strong augmentation}
\setlength{\tabcolsep}{4pt}
\begin{tabular}{c | c c c c c | c c c c c} 
 \hline \hline
 \multirow{2}{*}{Method} & \multicolumn{5}{c|}{Dog} & \multicolumn{5}{c}{Sheep} \\
 \cline{2-11}
  & Eye & Hoof & Knee & Elb & All & Eye & Hoof & Knee & Elb & All \\ 
 
 \hline
 Source only & 39.8 & 22.8 & 16.5 & 17.4 & 22.0 & 42.6 & 31.0 & 28.2 & 21.4 & 29.3\\ 
 Source only$^*$ & 26.6 & 44.0 & 30.8 & 25.1 & 32.2 & 53.3 & 63.0 & 51.5 & 32.1 & 49.6\\ 
 Oracle & 88.8 & 74.9 & 57.1 & 51.1 & 65.1 & 88.2 & 84.9 & 79.9 & 59.6 & 76.9\\ 
 \hline
 CCSSL~\cite{MuQiHaYu20}  & 24.7 & 37.4 & 25.4 & 19.6 & 27.0 & 44.3 & 55.4 & 43.5 & 28.5 & 42.8\\

 UDA-Animal~\cite{LiLe21} & 26.2 & 39.8 & 31.6 & 24.7 & 31.1 & 48.2 & 52.9 & 49.9 & 29.7 & 44.9\\

 RegDA~\cite{JiangJiWaLi0aLo21} & 46.8 & 54.6 & 32.9 & 31.2 & 40.6 & \textbf{62.8} & 68.5 & 57.0 & 42.4 & 56.9\\ 
 \hline
 Ours & \textbf{56.1} & \textbf{59.2} & \textbf{38.9} & \textbf{32.7} & \textbf{45.4} & 61.6 & \textbf{77.4} & \textbf{57.7} & \textbf{44.6} & \textbf{60.2}\\ 
 \hline
\end{tabular} 
\label{tab:animal_other}
\end{table}

\subsection{Generalization to Unseen Domains and Objects}
\label{exp:generalization}

So far, we have focused on accuracy in a given target domain, but we may face other types of unseen domains during training in real-world applications. Thus, we compare the generalization capacity of our method with baselines in a domain generalization setting where we test models on unseen domains and objects. 

\noindent\textbf{Domain Generalization on FreiHAND.} For hand pose estimation, we test models adapted on the RHD$\rightarrow$H3D setting with the other real-world hand dataset FreiHAND (FHD). We compare the accuracy on FHD and measure how well each method generalizes on the unseen domain FHD. As presented in Table~\ref{tab:gen}, the test performance on FHD is generally poor compared to the source only and oracle performance, presumably because of the larger domain gap between H3D and FHD. It is worth noticing the performance of CCSSL is lower than the source-only, even if it outperforms that in the \textsl{RHD$\rightarrow$H3D} setting by a large margin, revealing its lack of generalization capacity to the unseen domain, probably because of the lack of input-level alignment. On the other hand, RegDA and our method show better ability to generalize while ours achieves the best performance under most circumstances. 

\begin{table}[t]
\centering
\caption{  Domain generalization experiments on FreiHand (FHD) and Human3.6M. We report PCK@0.05. Fin: Fingertip. Sld: shoulder, Elb: Elbow. Source only indicates training only on RHD or SURREAL while Oracle indicates training only on FHD or Human3.6M}

\setlength{\tabcolsep}{4pt}
\resizebox{\textwidth}{!}{\begin{tabular}{c | c c c c c | c c c c c c c} 
 \hline \hline
 
 \multirow{2}{*}{Method} & \multicolumn{5}{c|}{\textsl{FreiHand}} & \multicolumn{7}{c}{\textsl{Human3.6M}} \\
 \cline{2-13}
  & MCP & PIP & DIP & Fin & All & Sld & Elb & Wrist & Hip & Knee & Ankle & All \\ 
 \hline
 Source only & 34.9 & 48.7 & 52.4 & 48.5  & 45.8 & 51.5 & 65.0 & 62.9 & 68.0 & 68.7 & 67.4 & 63.9 \\ 

 Oracle  & 92.8 & 90.3 & 87.7 & 78.5 & 87.2 & 95.3 & 91.8 & 86.9 & 95.6 & 94.1 & 93.6 & 92.9 \\ 
 \hline
 CCSSL~\cite{MuQiHaYu20} & 34.3 & 46.3 & 48.4 & 44.4 & 42.6 & 52.7 & 76.9 & 63.1 & 31.6 & 75.7 & 72.9 & 62.2 \\

 UDA-Animal~\cite{LiLe21}  & 29.6 & 46.6 & 50.0 & 45.3 & 42.2 & 54.4 & 75.3 & 62.1 & 21.6 & 70.4 & 69.2 & 58.8 \\

 RegDA~\cite{JiangJiWaLi0aLo21} & \textbf{37.8} & 51.8 & 53.2 & 47.5 & 46.9 & 76.9 & 80.2 & 69.7 & \textbf{52.0} & 80.3 & 80.0 & 73.2\\ 
 \hline
 Ours  & 35.6 & \textbf{52.3} & \textbf{55.4} & \textbf{50.6} & \textbf{47.1} & \textbf{77.0} & \textbf{85.9} & \textbf{73.8} & 47.6 & \textbf{80.7} & \textbf{80.6} & \textbf{74.3}\\ 
 \hline
\end{tabular} }
\label{tab:gen}
\end{table}
\noindent\textbf{Domain Generalization on Human3.6M.} We test the generalization ability of a model adapted from \textsl{SURREAL$\rightarrow$LSP} on Human3.6M. It should be noted that LSP contains only 2K images which are very small compared to Human3.6M. Thus, this task is challenging since we use small number of real data for domain generalization. In Table.~\ref{tab:gen}, we show that our method can generalize better than the baselines and achieves 74.3\% of accuracy. Our accuracy on the generalization task (74.3\%) is also comparable to the baselines performances of \textsl{SURREAL$\rightarrow$Human3.6M} (\eg, RegDA: 75.6), by using only $2k$ images. 

\begin{SCtable}[][t]
\centering
\caption{ Domain generalization experiments on AnimalPose. We report PCK@0.05. Source only indicates training only on Synthetic Animal}

\setlength{\tabcolsep}{4pt}
\resizebox{0.55\textwidth}{!}{\begin{tabular}{c | c c c c c c c} 
 \hline \hline
 Method & Horse & Dog & Cat & Sheep & Cow & All \\ 
 \hline
 Source only & 52.2 & 31.0 & 14.7 & 37.5 & 41.8 & 33.4 \\ 

 \hline
 CCSSL~\cite{MuQiHaYu20}  & 59.8 & 31.1 & 16.6 & 46.4 & 48.9 & 37.7 \\

 UDA-Animal~\cite{LiLe21} & \textbf{63.2} & 32.4 & 17.6 & 48.3 & 53.0 & 39.8\\

 RegDA~\cite{JiangJiWaLi0aLo21} & 58.4 & 34.9 & 17.4 & 45.1 & 46.3 & 39.0\\ 
 \hline
 Ours & 61.6 & \textbf{40.7} & \textbf{21.6} & \textbf{50.1} & \textbf{53.5} & \textbf{44.0}\\ 
 \hline
\end{tabular} }
\label{tab:animal_gen}
\end{SCtable}

\noindent\textbf{Domain Generalization on AnimalPose.} Finally, we evaluate the generalization capacity of models adapted from \textsl{SynAnimal$\rightarrow$TigDog} and test it on Animal Pose Dataset. It should be noted that models are only trained on horse and tiger images from the Synthetic Animal Dataset and tested on unseen animals (\eg, dog)  in Animal Pose Dataset.
Based on the results in Table~\ref{tab:animal_gen}, we can also witness an obvious improvement of our method above all the baselines and generalize better on unseen animals from unseen domains.

\begin{figure}[t]
\centering
    \begin{subfigure}[t]{0.19\textwidth}
        \includegraphics[width=0.99\linewidth]{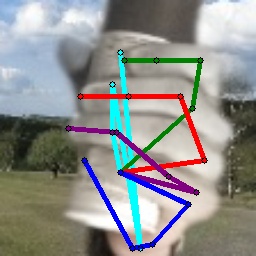}   
    \end{subfigure}
    \begin{subfigure}[t]{0.19\textwidth}
        \includegraphics[width=0.99\linewidth]{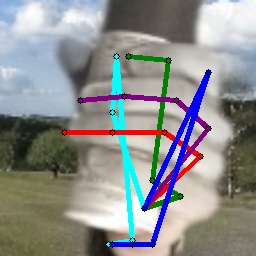}  
    \end{subfigure}
    \begin{subfigure}[t]{0.19\textwidth}
        \includegraphics[width=0.99\linewidth]{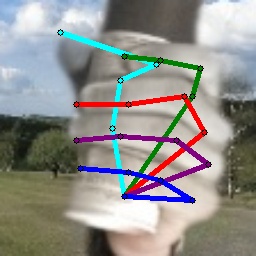}  
    \end{subfigure}
    \begin{subfigure}[t]{0.19\textwidth}
        \includegraphics[width=0.99\linewidth]{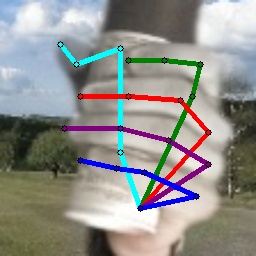}  
    \end{subfigure}
    \begin{subfigure}[t]{0.19\textwidth}
        \includegraphics[width=0.99\linewidth]{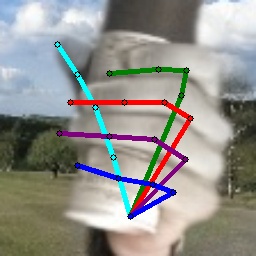}  
    \end{subfigure}
    \par\medskip
    \begin{subfigure}[t]{0.19\textwidth}
        \includegraphics[width=0.99\linewidth]{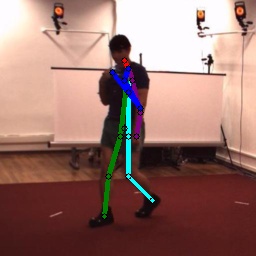}   
        \captionsetup{labelformat=empty}
        \caption{CCSSL}
    \end{subfigure}
    \begin{subfigure}[t]{0.19\textwidth}
        \includegraphics[width=0.99\linewidth]{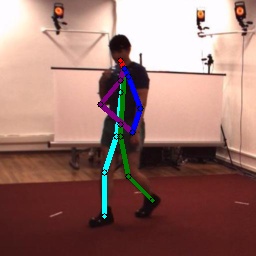}  
        \captionsetup{labelformat=empty}
        \caption{UDA-Animal}
    \end{subfigure}
    \begin{subfigure}[t]{0.19\textwidth}
        \includegraphics[width=0.99\linewidth]{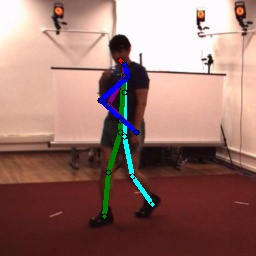}  
        \captionsetup{labelformat=empty}
        \caption{RegDA}
    \end{subfigure}
    \begin{subfigure}[t]{0.19\textwidth}
        \includegraphics[width=0.99\linewidth]{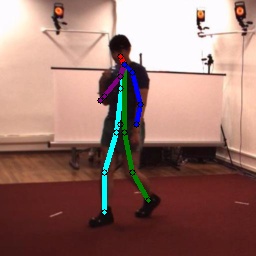}  
        \captionsetup{labelformat=empty}
        \caption{Ours}
    \end{subfigure}
    \begin{subfigure}[t]{0.19\textwidth}
        \includegraphics[width=0.99\linewidth]{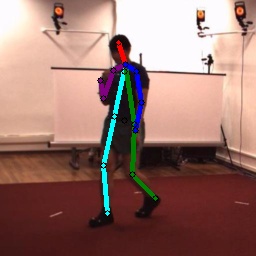}  
        \captionsetup{labelformat=empty}
        \caption{Ground-Truth}
    \end{subfigure}
    \caption{ Qualitative results of generalization to unseen domains}
    \label{fig:gen}
\end{figure}

\noindent\textbf{Qualitative Results.} We provide additional qualitative results on generalization in Figs.~\ref{fig:gen}. In Fig.~\ref{fig:gen}, it is clear that the baselines proposed for animal pose estimation do not work well. Our method produces more accurate keypoints compared to baselines. More qualitative results on animal are available in the appendix.

\begin{figure}[t]
\centering
    \begin{subfigure}[t]{0.3\textwidth}
        \includegraphics[width=0.99\linewidth]{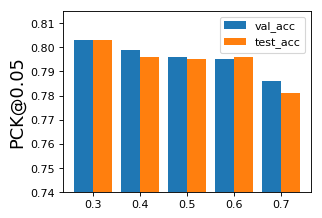}  
        \caption{Thresholding Ratio}
    \end{subfigure}
    \begin{subfigure}[t]{0.3\textwidth}
        \includegraphics[width=0.99\linewidth]{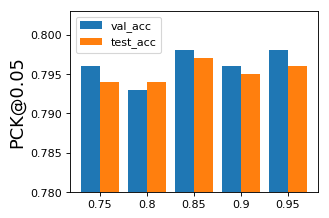}
        \caption{Occlusion Threshold}
    \end{subfigure}
    \begin{subfigure}[t]{0.3\textwidth}
        \includegraphics[width=0.99\linewidth]{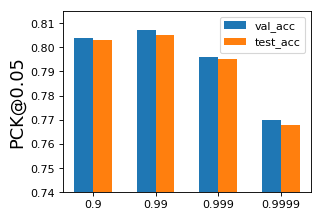}
        \caption{Teacher Momentum}
    \end{subfigure}
    \caption{ Sensitivity analysis on the thresholding, occlusion ratio, and momentum. Our method shows stable performance over hyper-parameters}
    \label{fig:sen}
\end{figure}

\subsection{Sensitivity Analysis}
\label{exp:sensitivity_analysis}
To further validate the robustness and generalization capacity of our method, we conducted sensitivity analysis regarding three major hyper-parameters in our framework, including the confidence thresholding ratio $p$, occlusion thresholding value $\tau_{occ}$, the momentum $\eta$ in Mean Teacher on \textsl{RHD$\rightarrow$H3D}. Additionally, we randomly split a separate validation set with the same size as the test set
from the target domain training data to simulate the hyper-parameter tuning process and avoid directly tuning the test accuracy. Based on the results presented in Fig.~\ref{fig:sen}, we find that our framework works stably under various settings. Meanwhile, we also find that the performance gradually decreases when we have a higher thresholding ratio for pseudo-labels, presumably because it brings in lower confident predictions as pseudo-labels and that deteriorates the unsupervised learning process. Also, we find that a greater teacher momentum is more likely to limit the framework to learn actively and harm the performance. 
More importantly, we can also learn that the validation accuracy
in all experiments is highly correlated with that on the test sets, which also indicates the generalization capacity of our method and the reliability to give indicative clues when tuning hyper-parameters on a separate validation set.

\subsection{Ablation Studies}
\label{exp:ablation}
We perform ablation studies in our framework to test their effectiveness and interaction with the rest of the framework. This also justify our other motivations 
regarding the task and the framework. Experiments are conducted under our major benchmarks including  \textsl{RHD$\rightarrow$H3D} and \textsl{SynAnimal$\rightarrow$TigDog}. Additional ablation studies can be found in the appendix.

Based on Table~\ref{tab:ab}, our framework can benefit from the heatmap normalization (denoted by Norm) that stabilizes the drifting effect and enables effective unsupervised learning from pseudo-labels via output-level domain alignment. Nevertheless, experiments on animal adaptation tasks show that such alignment might not be sufficiently helpful. Instead, more improvements are brought by the style transfer module, which confirms our reasoning that input-level variance is the major challenge in this task and can be mitigated by input-level alignments.

Adaptive occlusion can also provide extra focus on learning to detect occluded keypoints, as we can observe from \textsl{RHD$\rightarrow$H3D}. However such improvements are not reflected in \textsl{SynAnimal$\rightarrow$TigDog}. Considering the qualitative results in Figs.~\ref{fig:fig-1}, we conjecture that it is because the improvements in detecting occluded keypoints are not verifiable as their annotations are not available in the real animal dataset and therefore these predictions are not included in the PCK@0.05 evaluation protocol. More ablation studies are available in the appendix.

\begin{table}[t]
\centering
\caption{Ablation studies on hand \& animal pose estimation. Fin: Fingertip. MT: Mean Teacher, Norm: Heatmap Normalization, Style: Stylization, Occ: Adapt.\ Occlusion}
\setlength{\tabcolsep}{4pt}
\resizebox{\textwidth}{!}{\begin{tabular}{c | c c c c c | c c c c c c c c } 
 \hline \hline
 \multirow{2}{*}{Method} & \multicolumn{5}{c|}{\textsl{RHD$\rightarrow$H3D}} & \multicolumn{8}{c}{\textsl{SynAnimal$\rightarrow$TigDog}} \\
 \cline{2-14}
 & MCP & PIP & DIP & Fin & All & Eye & Chin & Sld & Hip & Elb & Knee & Hoof & All \\  
 \hline
 MT & 83.5 & 81.2 & 74.6 & 67.3 & 76.9 & 92.8 & 89.2 & 57.7 & 73.5 & 61.3 & 58.6 & 66.1 & 67.0\\ 
 MT + Norm  & 86.1 & 84.4 & 77.2 & 67.2 & 78.8& 91.9 & 89.9 & 59.3 & 62.7 & 60.8 & 67.6 & 64.1 & 68.1\\
 MT + Style & 84.6 & 82.5 & 76.6 & 66.9 & 77.6 & 95.0 & 93.8 & 57.8 & 74.7 & 63.5 & 67.4 & 67.4 & 70.4 \\

 MT + Norm + Style  & 86.6 & 84.4 & 78.3 & 68.1 & 79.1& 95.9 & 94.7 & 65.7 & 68.2 & 64.9 & 71.7 & 72.3 & 73.4\\

 MT + Norm + Style + Occ  & 86.7 & 84.6 & 78.9 & 68.1 & 79.6& 95.7 & 94.7 & 64.1 & 69.0 & 64.5 & 70.7 & 69.8 & 72.4 \\ 
 \hline
\end{tabular} }
\label{tab:ab}
\end{table}

\section{Conclusion}

While existing baselines focus on specific scenarios, we propose a unified framework that can be applied to diverse problems of domain adaptive pose estimation including hand pose, human body, and animal pose estimation.  Considering the challenges from different types of domain shifts, our method addresses both input and output-level discrepancies across domains and enables a more generic adaptation paradigm.
Extensive experiments demonstrate that our method not only achieves state-of-the-art performance under various domain adaptation scenarios but also exhibits excellent generalization capacity to unseen domains and objects. We hope our work can unify branches from different directions and provide a solid baseline for following works in this line of research.\newline

\noindent \textbf{Acknowledgements.} This work has been partially supported by NSF Award, DARPA, DARPA LwLL, ONR MURI grant N00014-19-1-2571 associated with AUSMURIB000001 (to M.B.) and by NSF grant 1535797, 1551572, (to. M.B.).

\clearpage

\bibliographystyle{splncs04}
\bibliography{egbib}

\clearpage

\section{Appendix}

In this supplementary material, we provide additional training details of our method. In addition to the ablation studies in the main paper, we also provide additional ablation studies on the \textsl{SURREAL$\rightarrow$Human3.6M} dataset. Finally, we show additional qualitative examples.

\subsection{Additional Training Details}
\begin{figure}[h]
\centering
    \includegraphics[width=0.9\linewidth]{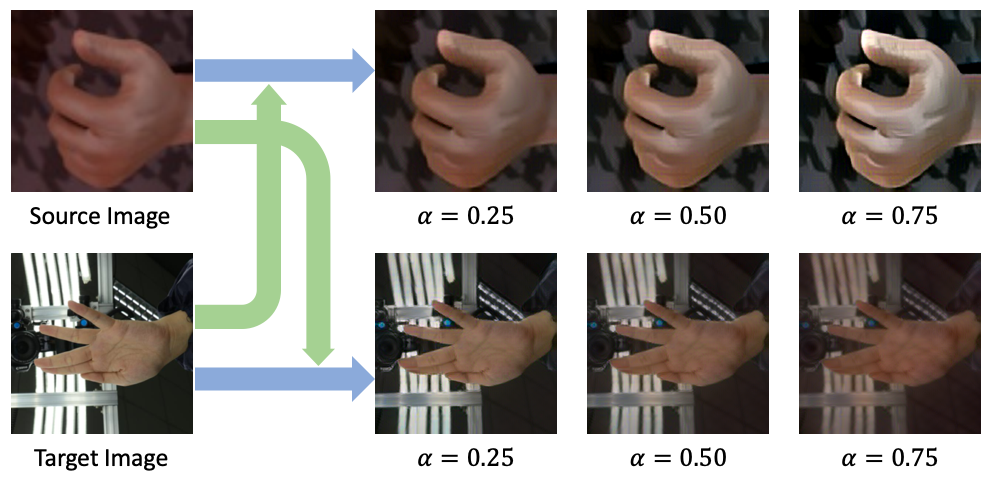}   
    \caption{ An illustration of style transfer between source and target domains with different content-style trade-off parameter $\alpha$. Blue arrows: content. Green arrows: style}
    \label{fig:style}
\end{figure}

We follow settings from AdaIN to train the generator $g$ from Eq. 2 with a content loss and a style loss balanced by style loss weight $\lambda=0.1$, on images with a resolution of $256 \times 256$. Exemplar results are illustrated in Fig.~\ref{fig:style}. 
During the training process of our framework, the pre-trained style transfer module will be fixed and perform bidirectional style transfer with a probability of $0.5$ in both our supervised and unsupervised learning branch with the content-style trade-off parameter $\alpha$ uniformly sampled from 0 to 1.  

Our pose estimation model $h$ is trained with input images with a resolution of $256\times 256$ and output heatmaps with a size of $64\times 64$, with the batch size of $32$ in each iteration, following our baselines~\cite{JiangJiWaLi0aLo21}. 

As for our adaptive keypoint occlusion, we randomly select keypoints with maximum activation greater than the occlusion threshold $\tau_{occ}$ and occlude it with a probability of $0.5$. The keypoints will be occluded by a patch from a random position in the same image with the size of $20 \times 20$.

\subsection{Additional Ablation Studies}

\begin{table}[h]
\centering
\caption{Ablation studies on  \textsl{SURREAL$\rightarrow$Human3.6M}. Sld: shoulder, Elb: Elbow. MT: Mean Teacher, Norm: Heatmap Normalization, Style: Stylization, Occ: Adaptive Occlusion}

\begin{tabular}{c | c c c c c c c} 
 \hline \hline
 Method & Sld & Elb & Wrist & Hip & Knee & Ankle & All \\ 
 \hline
 MT & 69.8 & 86.7 & 75.4 & 27.5 & 80.9 & 83.6 & 70.6\\ 
 MT + Norm  & 76.7 & 88.6 & 80.3 & 50.6 & 85.2 & 85.8 & 77.9\\
 MT + Style & 74.8 & 88.7 & 79.3 & 40.1 & 83.5 & 85.7 & 75.4 \\

 MT + Norm + Style  & 75.0 & 88.2 & 79.2 & 49.1 & 83.8 & 85.9 & 76.8\\

 MT + Norm + Style + Occ  & 78.1 & 89.6 & 81.1 & 52.6 & 85.3 & 87.1 & 79.0 \\ 
 \hline
\end{tabular} 
\label{tab:ab_s2h}
\end{table}

\begin{table}[h]
\centering
\caption{Ablation studies on data augmentation }
\begin{tabular}{c c c c c | c } 
 \hline
 Translation & Scale & Color & Rotation & Shear & PCK@0.05 \\ 
 \hline
\checkmark & & & & & 53.2\\
 & \checkmark & & & & 54.2\\
 &  & \checkmark & & & 51.7\\
 &  &  & \checkmark & & 77.7\\
 &  &  &  & \checkmark & 54.8\\
 \hline
\checkmark & \checkmark & & & & 54.4\\
\checkmark & \checkmark & \checkmark & & & 54.7\\
\checkmark & \checkmark & \checkmark & \checkmark & & 79.1\\
\checkmark & \checkmark & \checkmark & \checkmark & \checkmark & 79.6\\
 \hline
\end{tabular}
\label{tab:ab_aug}
\end{table}

\begin{figure}[h!]
\centering
    \begin{subfigure}[t]{0.19\textwidth}
        \includegraphics[width=0.99\linewidth]{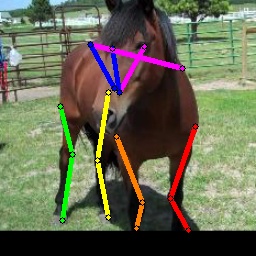}   
    \end{subfigure}
    \begin{subfigure}[t]{0.19\textwidth}
        \includegraphics[width=0.99\linewidth]{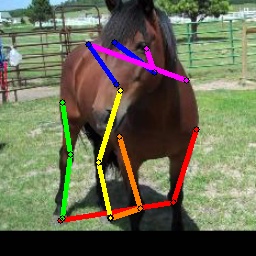}  
    \end{subfigure}
    \begin{subfigure}[t]{0.19\textwidth}
        \includegraphics[width=0.99\linewidth]{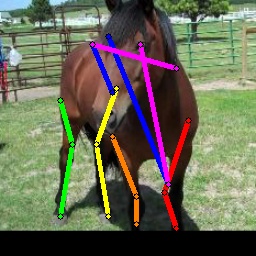}  
    \end{subfigure}
    \begin{subfigure}[t]{0.19\textwidth}
        \includegraphics[width=0.99\linewidth]{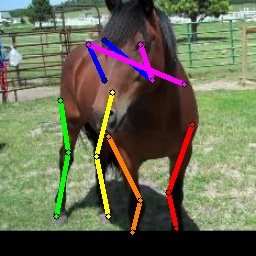}  
    \end{subfigure}
    \begin{subfigure}[t]{0.19\textwidth}
        \includegraphics[width=0.99\linewidth]{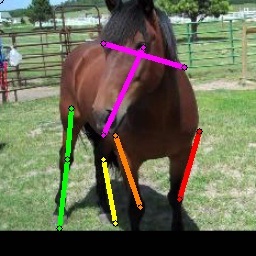}  
    \end{subfigure}
    \par\medskip
    \begin{subfigure}[t]{0.19\textwidth}
        \includegraphics[width=0.99\linewidth]{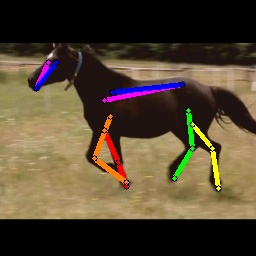}   
    \end{subfigure}
    \begin{subfigure}[t]{0.19\textwidth}
        \includegraphics[width=0.99\linewidth]{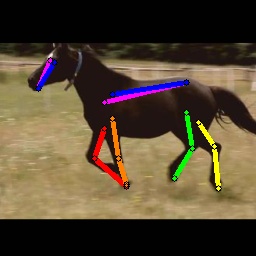}  
    \end{subfigure}
    \begin{subfigure}[t]{0.19\textwidth}
        \includegraphics[width=0.99\linewidth]{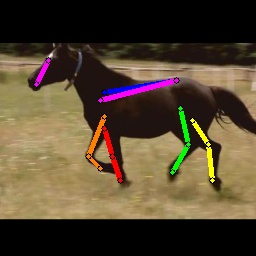}  
    \end{subfigure}
    \begin{subfigure}[t]{0.19\textwidth}
        \includegraphics[width=0.99\linewidth]{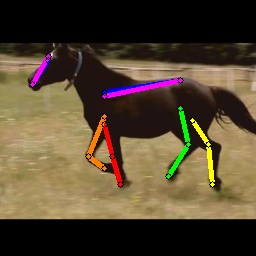}  
    \end{subfigure}
    \begin{subfigure}[t]{0.19\textwidth}
        \includegraphics[width=0.99\linewidth]{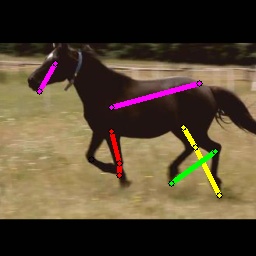}  
    \end{subfigure}
    \par\medskip
    \begin{subfigure}[t]{0.19\textwidth}
        \includegraphics[width=0.99\linewidth]{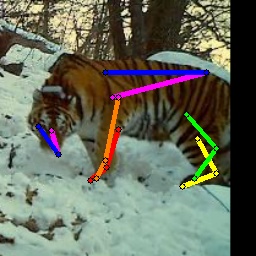}   
    \end{subfigure}
    \begin{subfigure}[t]{0.19\textwidth}
        \includegraphics[width=0.99\linewidth]{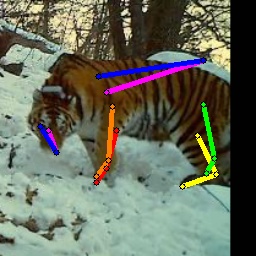}  
    \end{subfigure}
    \begin{subfigure}[t]{0.19\textwidth}
        \includegraphics[width=0.99\linewidth]{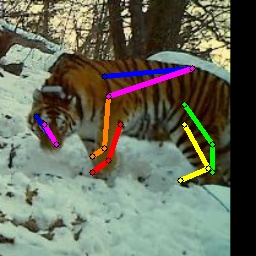}  
    \end{subfigure}
    \begin{subfigure}[t]{0.19\textwidth}
        \includegraphics[width=0.99\linewidth]{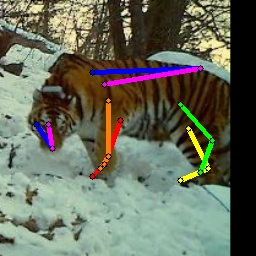}  
    \end{subfigure}
    \begin{subfigure}[t]{0.19\textwidth}
        \includegraphics[width=0.99\linewidth]{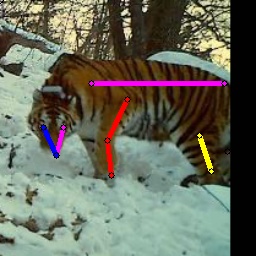}  
    \end{subfigure}
    \par\medskip
    \begin{subfigure}[t]{0.19\textwidth}
        \includegraphics[width=0.99\linewidth]{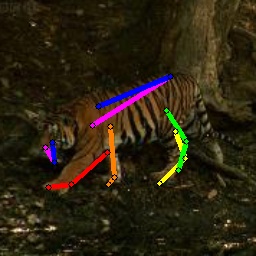}   
        \captionsetup{labelformat=empty}
        \caption{CCSSL}
    \end{subfigure}
    \begin{subfigure}[t]{0.19\textwidth}
        \includegraphics[width=0.99\linewidth]{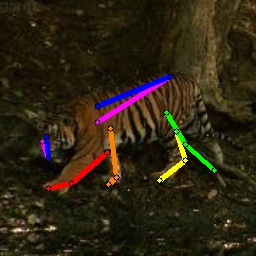}  
        \captionsetup{labelformat=empty}
        \caption{UDA-Animal}
    \end{subfigure}
    \begin{subfigure}[t]{0.19\textwidth}
        \includegraphics[width=0.99\linewidth]{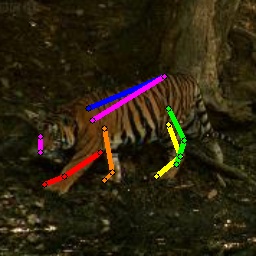}  
        \captionsetup{labelformat=empty}
        \caption{RegDA}
    \end{subfigure}
    \begin{subfigure}[t]{0.19\textwidth}
        \includegraphics[width=0.99\linewidth]{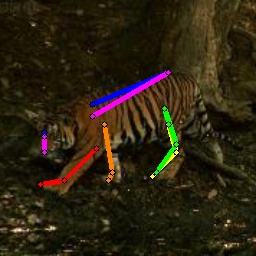}  
        \captionsetup{labelformat=empty}
        \caption{Ours}
    \end{subfigure}
    \begin{subfigure}[t]{0.19\textwidth}
        \includegraphics[width=0.99\linewidth]{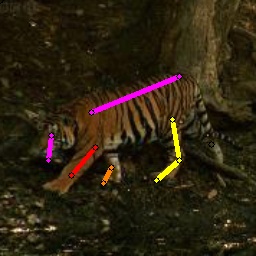}  
        \captionsetup{labelformat=empty}
        \caption{Ground-Truth}
    \end{subfigure}
    \caption{Additional qualitative results on TigDog. Compared with baselines, our method can more accurately estimate the position of keypoints from different angle (the first row), different motion (the second row), and different animals. It is also worth noting that the position of the legs in the example at the second row is mistakenly annotated in ground-truth, while we can still estimate their actual position. This justifies the motivation of our work that seeks to free pose estimation tasks from the dependence of the laborious and unreliable manual annotation process}
    \label{fig:supp_qual_1}
\end{figure}

In addition to \textsl{RHD$\rightarrow$H3D} and \textsl{SynAnimal$\rightarrow$TigDog}, we also present ablation studies on another major benchmark, \textsl{SURREAL$\rightarrow$Human3.6M} in Table~\ref{tab:ab_s2h}. Based on the results we can observe a greater improvement after applying heatmap normalization (the first and the second row), showing the necessity of addressing the drift effect under this scenario. On the other hand, we can also observe fewer improvements (the third and the fourth row) brought by the style transfer module, which coincide with our conclusion from the ablation studies on \textsl{RHD$\rightarrow$H3D} that the major challenge in human pose estimation tasks comes from the output-level discrepancy instead of the input-level. On that basis, our adaptive keypoint occlusion mechanism further boosts the performance by 2.2 percent points (the last row) and achieves the state-of-the-art performance, which shows the effectiveness of the occlusion mechanism specialized in this task.

\subsection{Ablation studies of data augmentation}

Tab.~\ref{tab:ab_aug} presents ablation studies of data augmentation methods on \textsl{RHD$\rightarrow$H3D}. We compare the performance of our method with different compositions of augmentations commonly used in pose estimation tasks, and we observe that rotation provides the most significant gain.

\subsection{Additional Qualitative Results \& Failure Cases}
\begin{figure}[h]
\centering
    \begin{subfigure}[t]{0.19\textwidth}
        \includegraphics[width=0.99\linewidth]{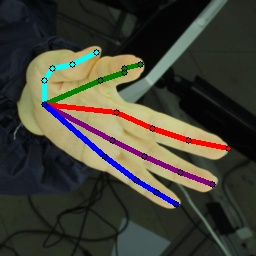}   
    \end{subfigure}
    \begin{subfigure}[t]{0.19\textwidth}
        \includegraphics[width=0.99\linewidth]{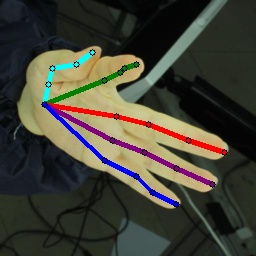}  
    \end{subfigure}
    \begin{subfigure}[t]{0.19\textwidth}
        \includegraphics[width=0.99\linewidth]{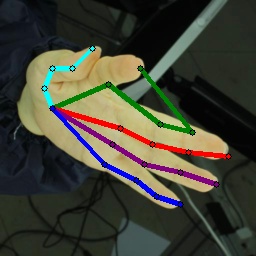}  
    \end{subfigure}
    \begin{subfigure}[t]{0.19\textwidth}
        \includegraphics[width=0.99\linewidth]{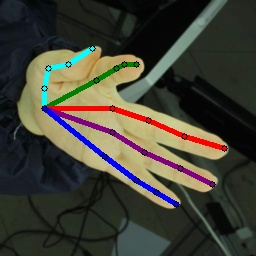}  
    \end{subfigure}
    \begin{subfigure}[t]{0.19\textwidth}
        \includegraphics[width=0.99\linewidth]{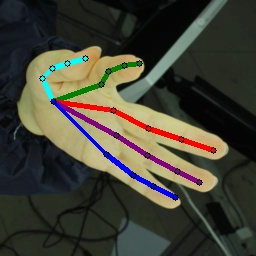}  
    \end{subfigure}
    \par\medskip
    \begin{subfigure}[t]{0.19\textwidth}
        \includegraphics[width=0.99\linewidth]{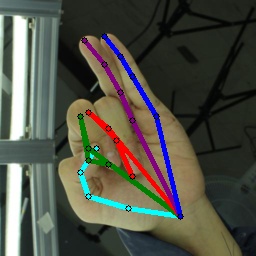}   
    \end{subfigure}
    \begin{subfigure}[t]{0.19\textwidth}
        \includegraphics[width=0.99\linewidth]{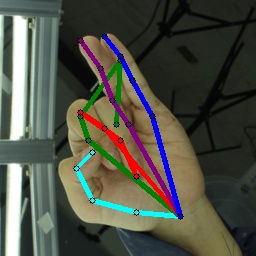}  
    \end{subfigure}
    \begin{subfigure}[t]{0.19\textwidth}
        \includegraphics[width=0.99\linewidth]{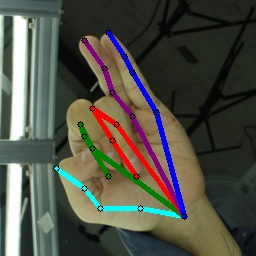}  
    \end{subfigure}
    \begin{subfigure}[t]{0.19\textwidth}
        \includegraphics[width=0.99\linewidth]{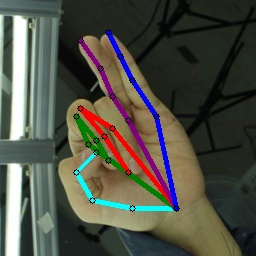}  
    \end{subfigure}
    \begin{subfigure}[t]{0.19\textwidth}
        \includegraphics[width=0.99\linewidth]{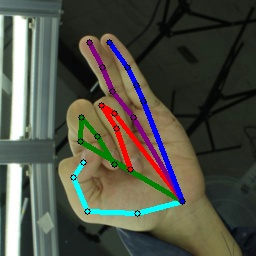}  
    \end{subfigure}
    \par\medskip
    \begin{subfigure}[t]{0.19\textwidth}
        \includegraphics[width=0.99\linewidth]{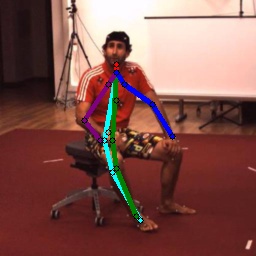}   
    \end{subfigure}
    \begin{subfigure}[t]{0.19\textwidth}
        \includegraphics[width=0.99\linewidth]{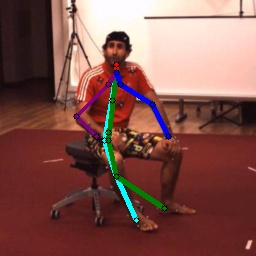}  
    \end{subfigure}
    \begin{subfigure}[t]{0.19\textwidth}
        \includegraphics[width=0.99\linewidth]{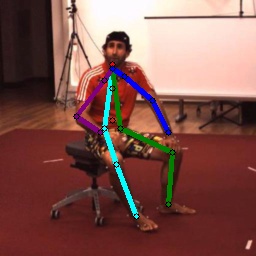}  
    \end{subfigure}
    \begin{subfigure}[t]{0.19\textwidth}
        \includegraphics[width=0.99\linewidth]{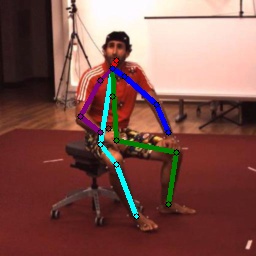}  
    \end{subfigure}
    \begin{subfigure}[t]{0.19\textwidth}
        \includegraphics[width=0.99\linewidth]{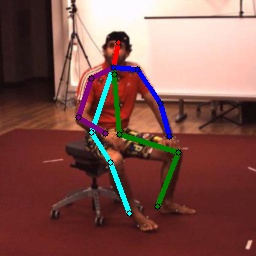}  
    \end{subfigure}
    \par\medskip
    \begin{subfigure}[t]{0.19\textwidth}
        \includegraphics[width=0.99\linewidth]{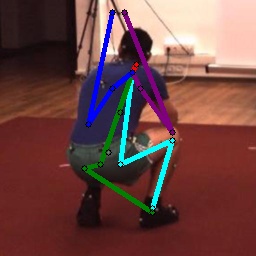}   
        \captionsetup{labelformat=empty}
        \caption{CCSSL}
    \end{subfigure}
    \begin{subfigure}[t]{0.19\textwidth}
        \includegraphics[width=0.99\linewidth]{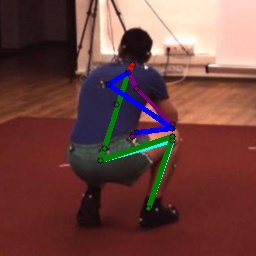}  
        \captionsetup{labelformat=empty}
        \caption{UDA-Animal}
    \end{subfigure}
    \begin{subfigure}[t]{0.19\textwidth}
        \includegraphics[width=0.99\linewidth]{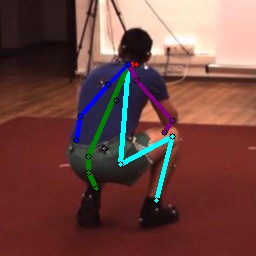}  
        \captionsetup{labelformat=empty}
        \caption{RegDA}
    \end{subfigure}
    \begin{subfigure}[t]{0.19\textwidth}
        \includegraphics[width=0.99\linewidth]{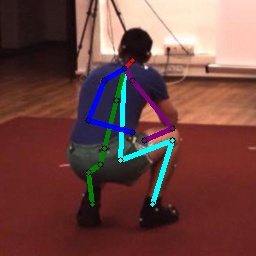}  
        \captionsetup{labelformat=empty}
        \caption{Ours}
    \end{subfigure}
    \begin{subfigure}[t]{0.19\textwidth}
        \includegraphics[width=0.99\linewidth]{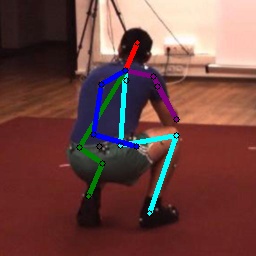}  
        \captionsetup{labelformat=empty}
        \caption{Ground-Truth}
    \end{subfigure}
    \caption{Additional qualitative results on human pose estimation tasks. We show that our method can better handle the detection of keypoints in diverse poses (the first and the second rows) and from diverse view points (the third and the fourth rows), compared with baselines}
    \label{fig:supp_qual_2}
\end{figure}

\begin{figure}[h!]
\centering
    \begin{subfigure}[t]{0.19\textwidth}
        \includegraphics[width=0.99\linewidth]{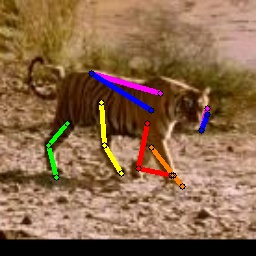}   
    \end{subfigure}
    \begin{subfigure}[t]{0.19\textwidth}
        \includegraphics[width=0.99\linewidth]{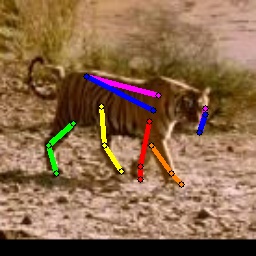}  
    \end{subfigure}
    \begin{subfigure}[t]{0.19\textwidth}
        \includegraphics[width=0.99\linewidth]{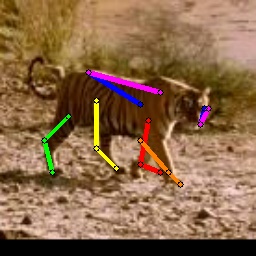}  
    \end{subfigure}
    \begin{subfigure}[t]{0.19\textwidth}
        \includegraphics[width=0.99\linewidth]{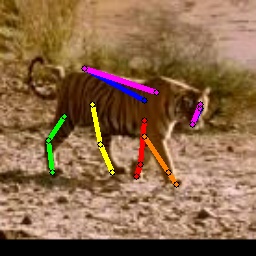}  
    \end{subfigure}
    \begin{subfigure}[t]{0.19\textwidth}
        \includegraphics[width=0.99\linewidth]{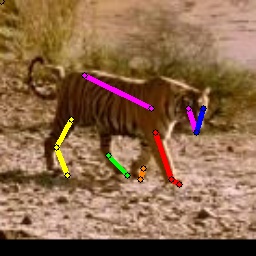}  
    \end{subfigure}
    \par\medskip
    \begin{subfigure}[t]{0.19\textwidth}
        \includegraphics[width=0.99\linewidth]{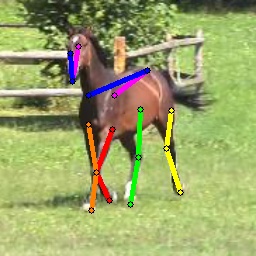}   
        \captionsetup{labelformat=empty}
        \caption{CCSSL}
    \end{subfigure}
    \begin{subfigure}[t]{0.19\textwidth}
        \includegraphics[width=0.99\linewidth]{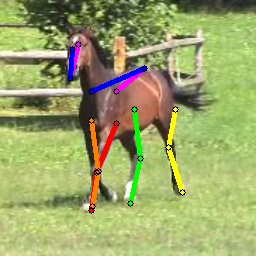}  
        \captionsetup{labelformat=empty}
        \caption{UDA-Animal}
    \end{subfigure}
    \begin{subfigure}[t]{0.19\textwidth}
        \includegraphics[width=0.99\linewidth]{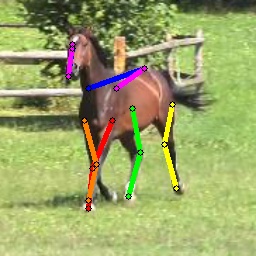}  
        \captionsetup{labelformat=empty}
        \caption{RegDA}
    \end{subfigure}
    \begin{subfigure}[t]{0.19\textwidth}
        \includegraphics[width=0.99\linewidth]{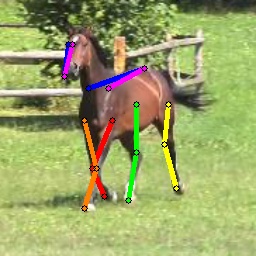}  
        \captionsetup{labelformat=empty}
        \caption{Ours}
    \end{subfigure}
    \begin{subfigure}[t]{0.19\textwidth}
        \includegraphics[width=0.99\linewidth]{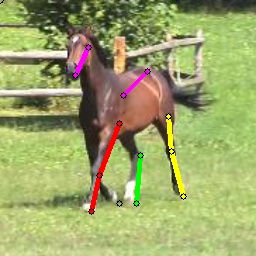}  
        \captionsetup{labelformat=empty}
        \caption{Ground-Truth}
    \end{subfigure}
    \caption{Failure cases on TigDog. We show that extreme cases in typical pose estimation problems, including distinguishing left and right limbs (the first row) and ambiguous occlusion (the second row), can still be challenges in our method and result in an incorrect prediction  }
    \label{fig:supp_qual_3}
\end{figure}

\begin{figure}[h!]
\centering
    \begin{subfigure}[t]{0.19\textwidth}
        \includegraphics[width=0.99\linewidth]{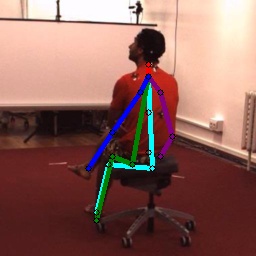}   
    \end{subfigure}
    \begin{subfigure}[t]{0.19\textwidth}
        \includegraphics[width=0.99\linewidth]{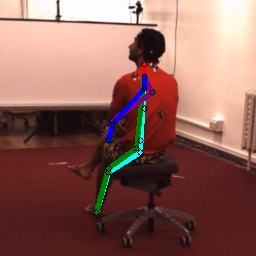}  
    \end{subfigure}
    \begin{subfigure}[t]{0.19\textwidth}
        \includegraphics[width=0.99\linewidth]{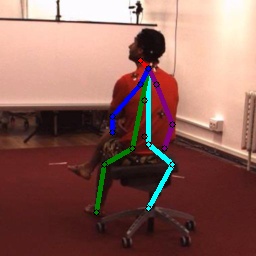}  
    \end{subfigure}
    \begin{subfigure}[t]{0.19\textwidth}
        \includegraphics[width=0.99\linewidth]{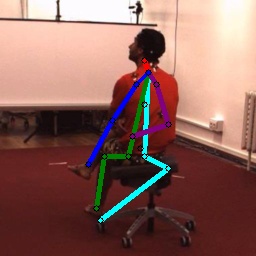}  
    \end{subfigure}
    \begin{subfigure}[t]{0.19\textwidth}
        \includegraphics[width=0.99\linewidth]{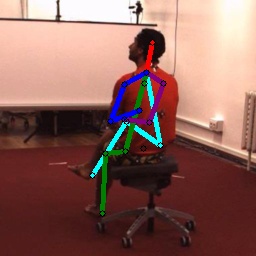}  
    \end{subfigure}
    \par\medskip
    \begin{subfigure}[t]{0.19\textwidth}
        \includegraphics[width=0.99\linewidth]{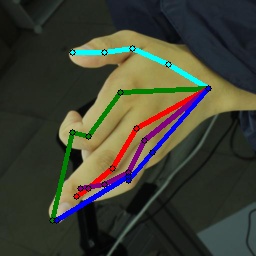}   
        \captionsetup{labelformat=empty}
        \caption{CCSSL}
    \end{subfigure}
    \begin{subfigure}[t]{0.19\textwidth}
        \includegraphics[width=0.99\linewidth]{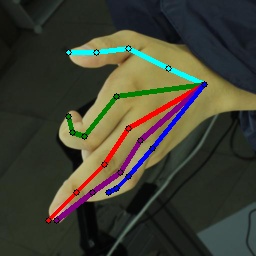}  
        \captionsetup{labelformat=empty}
        \caption{UDA-Animal}
    \end{subfigure}
    \begin{subfigure}[t]{0.19\textwidth}
        \includegraphics[width=0.99\linewidth]{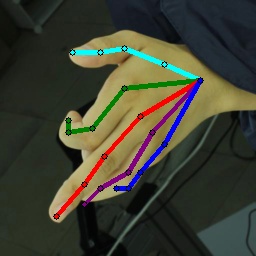}  
        \captionsetup{labelformat=empty}
        \caption{RegDA}
    \end{subfigure}
    \begin{subfigure}[t]{0.19\textwidth}
        \includegraphics[width=0.99\linewidth]{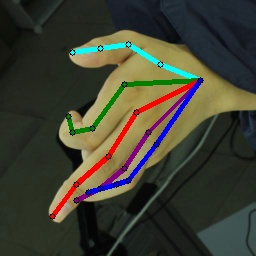}  
        \captionsetup{labelformat=empty}
        \caption{Ours}
    \end{subfigure}
    \begin{subfigure}[t]{0.19\textwidth}
        \includegraphics[width=0.99\linewidth]{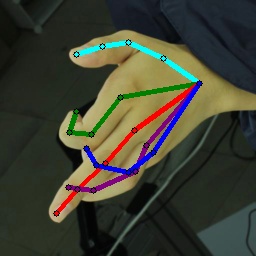}  
        \captionsetup{labelformat=empty}
        \caption{Ground-Truth}
    \end{subfigure}
    \caption{Failure cases on human pose estimation tasks. Existing difficulties in typical pose estimation tasks still pose a huge challenge to all the baseline methods and ours, especially when ambiguous occlusion happens  }
    \label{fig:supp_qual_4}
\end{figure}

\begin{figure}[h!]
\centering
    \begin{subfigure}[t]{0.19\textwidth}
        \includegraphics[width=0.99\linewidth]{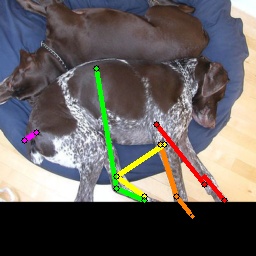}   
    \end{subfigure}
    \begin{subfigure}[t]{0.19\textwidth}
        \includegraphics[width=0.99\linewidth]{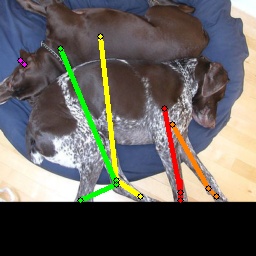}  
    \end{subfigure}
    \begin{subfigure}[t]{0.19\textwidth}
        \includegraphics[width=0.99\linewidth]{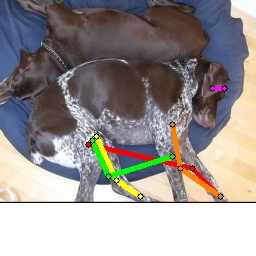}  
    \end{subfigure}
    \begin{subfigure}[t]{0.19\textwidth}
        \includegraphics[width=0.99\linewidth]{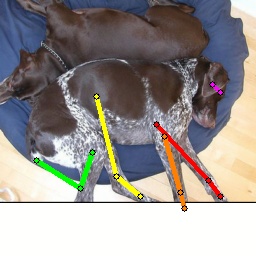}  
    \end{subfigure}
    \begin{subfigure}[t]{0.19\textwidth}
        \includegraphics[width=0.99\linewidth]{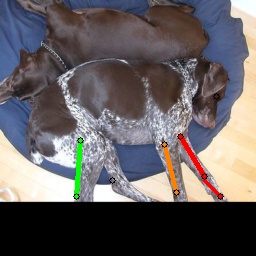}  
    \end{subfigure}
    \par\medskip
    \begin{subfigure}[t]{0.19\textwidth}
        \includegraphics[width=0.99\linewidth]{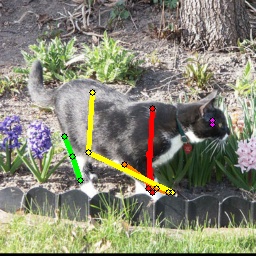}   
        \captionsetup{labelformat=empty}
        \caption{CCSSL}
    \end{subfigure}
    \begin{subfigure}[t]{0.19\textwidth}
        \includegraphics[width=0.99\linewidth]{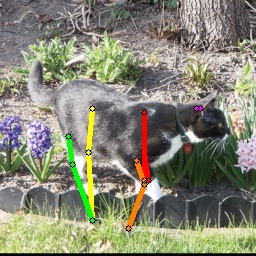}  
        \captionsetup{labelformat=empty}
        \caption{UDA-Animal}
    \end{subfigure}
    \begin{subfigure}[t]{0.19\textwidth}
        \includegraphics[width=0.99\linewidth]{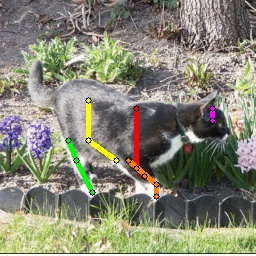}  
        \captionsetup{labelformat=empty}
        \caption{RegDA}
    \end{subfigure}
    \begin{subfigure}[t]{0.19\textwidth}
        \includegraphics[width=0.99\linewidth]{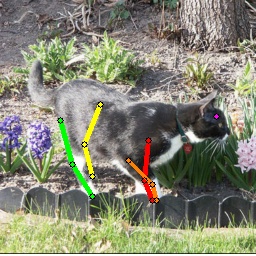}  
        \captionsetup{labelformat=empty}
        \caption{Ours}
    \end{subfigure}
    \begin{subfigure}[t]{0.19\textwidth}
        \includegraphics[width=0.99\linewidth]{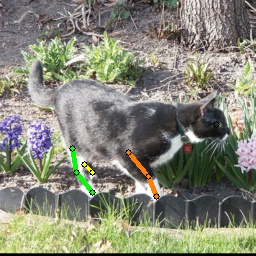}  
        \captionsetup{labelformat=empty}
        \caption{GT}
    \end{subfigure}
    \caption{ Qualitative results of generalization to unseen animals and domains. Note that the annotations for occluded keypoints (yellow parts) are not available in ground truth}
    \label{fig:gen_animal}
\end{figure}

We provide additional qualitative results in Figures.~\ref{fig:supp_qual_1}, ~\ref{fig:supp_qual_2},~\ref{fig:supp_qual_3},~\ref{fig:supp_qual_4}, and~\ref{fig:gen_animal}.

\section{Discussion of limitation \& future directions}
Even though our method gains significant improvements over source-only pre-training, the overall performance is still limited and not comparable to the supervised learning level (target only).
Therefore, while we explore only unsupervised DA, semi-supervised DA methods that can leverage a limited amount of target domain annotation to further improve the accuracy will be an interesting future direction. Additionally, while we focus on  domain adaptive 2D pose estimation, 3D pose estimation is also a good research direction to explore as it is harder to obtain depth annotations.
 
\section{Learning Animal pose estimation from human}
The main challenge under our fully unsupervised settings, if we learn only from a human pose dataset without animals, would be the limited number of shared keypoints because of the anatomical differences between human and animals, which limits the amount of information we can learn from the source human dataset. 
In SURREAL→Tigdog learning limbs of human and animals, our method achieves 7.9\% of accuracy, while the source-only pretraining and RegDA achieves 2.4\% and 2.6\% respectively. These low accuracies indicate the difficulty of the adaptation in this unsupervised manner when the anatomical differences are significant. 

\clearpage

\end{document}